\documentclass[final,5p,authoryear,a4paper]{elsarticle}

\usepackage{prletters}
\usepackage{palatino}
\usepackage{hyphenat}
\usepackage{hyperref}

\usepackage[utf8]{inputenc}
\usepackage{float}
\usepackage{amsmath}
\usepackage{apalike}
\DeclareMathOperator{\softmax}{softmax}
\DeclareMathOperator{\attention}{Attention}
\usepackage{graphicx}
\usepackage{caption}
\usepackage{subcaption}
\graphicspath{ {figures/} }

\begin{document}

\title{Transfer-learning for video classification: Video Swin Transformer on multiple domains}

\author[inesc,ist]{Daniel Oliveira}
\ead{daniel.oliveira@inesc-id.pt}
\author[inesc,ist]{David Martins de Matos}
\ead{david.matos@inesc-id.pt}

\address[inesc]{INESC-ID Lisboa, R. Alves Redol 9, 1000-029 Lisboa, Portugal}
\address[ist]{Instituto Superior Técnico, Universidade de Lisboa, Av. Rovisco Pais, 1049-001 Lisboa, Portugal}

\begin{abstract}
The computer vision community has seen a shift from convolutional-based to pure transformer architectures for both image and video tasks. Training a transformer from zero for these tasks usually requires a lot of data and computational resources. Video Swin Transformer (VST) is a pure-transformer model developed for video classification which achieves state-of-the-art results in accuracy and efficiency on several datasets. In this paper, we aim to understand if VST generalizes well enough to be used in an out-of-domain setting. We study the performance of VST on two large-scale datasets, namely FCVID and Something-Something using a transfer learning approach from Kinetics-400, which requires around 4x less memory than training from scratch. We then break down the results to understand where VST fails the most and in which scenarios the transfer-learning approach is viable. Our experiments show an 85\% top-1 accuracy on FCVID without retraining the whole model which is equal to the state-of-the-art for the dataset and a 21\% accuracy on Something-Something. The experiments also suggest that the performance of the VST decreases on average when the video duration increases which seems to be a consequence of a design choice of the model. From the results, we conclude that VST generalizes well enough to classify out-of-domain videos without retraining when the target classes are from the same type as the classes used to train the model. We observed this effect when we performed transfer-learning from Kinetics-400 to FCVID, where most datasets target mostly objects. On the other hand, if the classes are not from the same type, then the accuracy after the transfer-learning approach is expected to be poor. We observed this effect when we performed transfer-learning from Kinetics-400, where the classes represent mostly objects, to Something-Something, where the classes represent mostly actions.
\end{abstract}

\begin{keyword}
Video Classification, Action Classification, Transformers, Video Transformers, Video Swin Transformer, Transfer Learning, FCVID, Kinetics, Something-Something
\end{keyword}

\maketitle

\section{Introduction}
Recognizing and understanding the contents of images and videos is a crucial problem for many applications, such as image and video retrieval, smart advertising, allowing artificial agents to perceive the surrounding world, among others. Convolutional neural networks (CNN) have been widely used for video classification, namely 3D convolution \citep{3d_convolution} which is an extension of 2D convolution. Recently, we have seen a shift from convolution based architectures to transformer-based architectures. This shift started for image classification with the introduction of ViT \citep{vit}, a visual transformer for image classification. Later, \cite{vivit} proposed a pure transformer architecture for video classification that relied on the factorization of the spatial and temporal dimensions of the input. \cite{video_understanding} proposed the application of self-attention between space and time separately.
More recently, VST \citep{video_swin_transformer} proposed a pure transformer architecture for video classification that is able to surpass the factorized models in efficiency and accuracy by taking advantage of the spatiotemporal locality of videos.
Training a model to recognise images or videos from zero requires a lot of data and computational resources, namely memory. In this paper, we perform a study of the accuracy of the VST in a transfer-learning scenario which requires around 4x less memory than training from scratch. We aim to understand if Video Swin Transfer generalizes well enough to be used on an out of domain setting. We evaluate the performance of our approach on FCVID \citep{fcvid}, a large-scale in-the-wild dataset, and on Something-Something, a dataset of humans performing actions. Our approach takes the advantage of only having to train one layer to achieve the results stated.
This paper is organized as follows: in section~\ref{sec:video_swin_transormer} we describe the VST model. Section~\ref{sec:datasets}
 describes the datasets relevant for the experiments. Section~\ref{sec:experiments} describes our setup. Section~\ref{sec:conclusions} presents the conclusions from the experiments and proposals for future work.

\section{Video Swin Transformer}\label{sec:video_swin_transormer}
Video Swin Transformer (VST)~\citep{video_swin_transformer} is an adaptation of the Swin Transformer~\citep{swin_transformer} and currently achieves state-of-the-art results on the video classification task on Kinetics-400 \citep{kinetics_400}, Kinetics-600 \citep{kinetics_600}, and Something-Something \citep{something_something}.

\begin{figure*}[h!]
    \centering
    \includegraphics[width=1\linewidth]{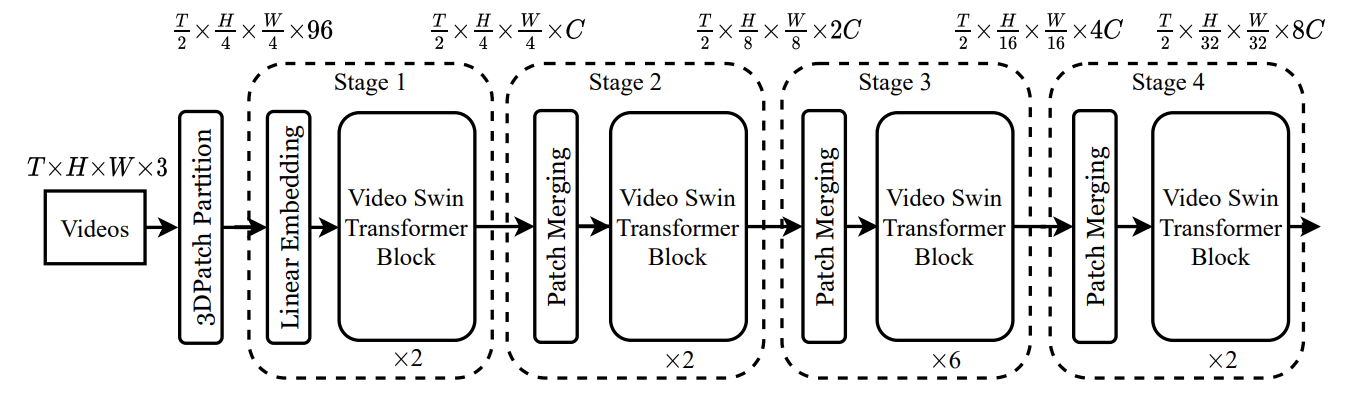}
    \caption{The architecture of VST \citep{vit}}
    \label{fig:video_swin_arquitecture}
\end{figure*}

The architecture of the model is shown in figure~\ref{fig:video_swin_arquitecture}. The input is a video, defined by a tensor with dimensions $T \times H \times W \times 3$, where T corresponds to the time dimension, i.e. the number of frames, and $H \times W \times 3$ to the number of pixels in each frame. VST applies 3-dimensional patching to the video \citep{vit}. Each patch has size $2 \times 4 \times 4 \times 3$. From patching the video results $\frac{T}{2} \times \frac{H}{4} \times \frac{W}{4}$ patches that represent the whole video. Each patch is represented by a C dimensional vector.

The Patch Merging layers placed between each VST Block on fig.~\ref{fig:video_swin_arquitecture} are responsible for merging groups of $2 \times 2$ patches and then applying a linear transformation that reduces the features to half of the original dimension.

\begin{figure}[h!]
    \centering
    \includegraphics[width=0.8\linewidth]{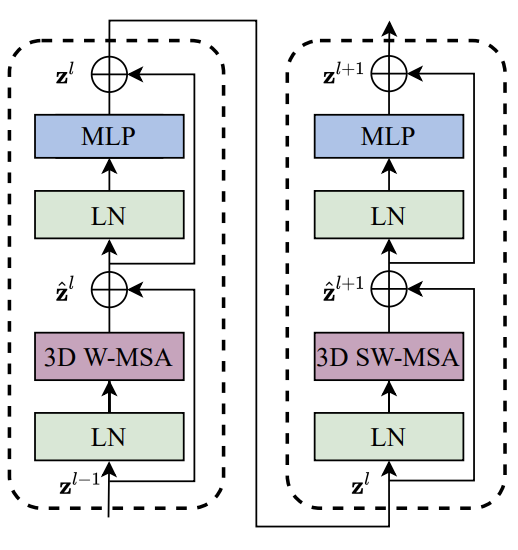}
    \caption{An illustration of two successive VST blocks. \citep{video_swin_transformer}}
    \label{fig:video_swin_transformer_block}
\end{figure}

The main block in this architecture is the VST block represented on Figure~\ref{fig:video_swin_transformer_block}, this block has the structure of a standard transformer where the Multi-Head self-attention module is replaced with a 3D-shifted window-based multi-head self-attention module.

Given a video composed of $T' \times H' \times W'$ patches and a 3D window with size $P \times M \times M$, the window is used to partition the video patches into $\frac{T'}{P} \times \frac{H'}{M} \times \frac{W'}{M}$ non-overlapping 3D windows.

After the first VST layer, the window partitioning configuration is shifted along the temporal, the height, and the width axis by $\left( \frac{P}{2}, \frac{M}{2}, \frac{M}{2}  \right)$ patches from the preceding layer. This shifted attention architecture was originally developed by \cite{swin_transformer} and introduces connections between the neighboring non-overlapping 3D windows from the previous layer.

A relative position bias is also added to each head in the self-attention blocks that was shown to be advantageous \citep{swin_transformer}. The attention computed by these blocks is expressed by equation~\ref{eq:bias_attention}. Where $Q, K, V \in R^{PM^{2} \times d}$ are the query, key, and value matrices. $d$ is the dimension of the features and $PM^2$ is the number of tokens in a 3D window.

\begin{equation}
\attention(Q,K,V) = \softmax\left(\frac{QK^T}{\sqrt{d_k}} + B\right)V
\label{eq:bias_attention}
\end{equation}

It is worth mentioning that in the last dense layer of the model just performs classification over the set of possible classes in the dataset used.

\section{Datasets}
\label{sec:datasets}
There are three datasets relevant to this work, namely Kinetics-400, FCVID and Something-Something, these datasets are described below.

\subsection{Kinetics}
Kinetics is a collection of three datasets, Kinetics-400 \citep{kinetics_400} Kinetics-600 \citep{kinetics_600} and Kinetics-700 \citep{kinetics_700}.
Each of these datasets contains a set of URLs to 10 second YouTube videos and one label describing the contents of the video. Kinetics covers 400/600/700 human action classes, depending on the version of the dataset. These classes include several human-to-object interactions such as steering a car, as well as human-to-human interactions, such as shaking hands.

\subsection{FCVID}
FCVID \citep{fcvid} is a large-scale in-the-wild dataset containing a total of 91,223 videos collected from YouTube with an average duration of 167 seconds. These videos are organized in 239 different categories such as social events, procedural events, objects and scenes. The videos were collected by performing YouTube searches using the categories as keywords. These categories were used as the labels for the videos and were revised by humans. Videos longer than 30 minutes were removed. Each video of FCVID can have one or more labels. One video for instance can be classified as "river" and "bridge" at the same time. Some classes appear to be sub classes of another class, such as dog/playingFrisbeeWithDog or river/waterfall,  despite the author's not providing any class hierarchy.

\subsection{Something-Something}
Something-Something \citep{something_something} is a collection of 220,847  videos with 174 different labels of humans performing actions with everyday objects. The videos created and labeled by 1,300 crowd workers. Each video ranges from 2 to 6 seconds. The labels are textual descriptions based on templates such as "Dropping
[something] into [something]” where "[something]" serves a placeholder for objects.
\\

The first two datasets, Kinetics-400 and FCVID cover different domains of videos with some overlapping in between. The task between these two datasets is similar, assigning the label that better describes the contents of the video. For the third dataset, Something-Something, the task is different since the objective is to assign the label that better describes the action being performed rather than the identifying the main object on the video.

The first dataset, Kinetics-400, is where our model was trained. FCVID and Something-Something, are the ones that we target using a transfer-learning approach. Our transfer-learning approach is explained in detail in the section\ref{sec:experiments}.

\section{Experiments}
\label{sec:experiments}

\subsection{Setup}
VST comes with 4 different model sizes, namely:
\begin{enumerate}
    \item Swin-T: C = 96, layers per block = (2, 2, 6, 2), parameters = 28M
    \item Swin-S: C = 96, layers per block = (2, 2, 18, 2), parameters = 50M
    \item Swin-B: C = 128, layers per block = (2, 2, 18, 2), parameters = 88M
    \item Swin-L: C = 192, layers per block = (2, 2, 18, 2), parameters = 88M
\end{enumerate}

For this experiment, we consider the smaller version, Swin-T, trained on Kinetics-400. We downloaded the model weights available on the official repository \footnote{\url{https://github.com/SwinTransformer/Video-Swin-Transformer} (visited on Oct/2022)}. 

The original model, trained for Kinetics-400, has a final dense layer with 400 neurons. For the FCVID experiment, we replaced this layer with a new dense layer containing 239 neurons, each neuron corresponds to an FCVID class. For our Something-Something experiment we replaced the final dense layer with one containing 174 neurons.
Similar to \cite{video_swin_transformer}, we also use a batch size of 64 and an AdamW \citep{adamw} optimizer for 30 epochs
using a cosine decay learning rate scheduler and 2.5 epochs of linear warm-up. We also sample a clip of 32 frames from each full-length video using a temporal stride of 2 and spatial size of $224 \times 224$ as done in the original paper. During the training phase we froze all the model layers except for the final layer. 

\subsection{Results}

\subsubsection{FCVID}
Our results on the test set of FCVID show an accuracy of 85.0\%. This value compares with an accuracy of the original model on Kinetics-400 of 78.8\%. This improvement in the accuracy can be explained by the number of classes 239 on FCVID vs 400 on Kinetics-400 which makes the problem easier for FCVID. This value is equal to the accuracy of AdaFocus V2 \citep{adamw} which achieves state-of-the-art results on FCVID. However, we have the advantage of training only one layer instead of training the full model.

\begin{figure}[h!]
    \centering
    \includegraphics[width=0.9\linewidth]{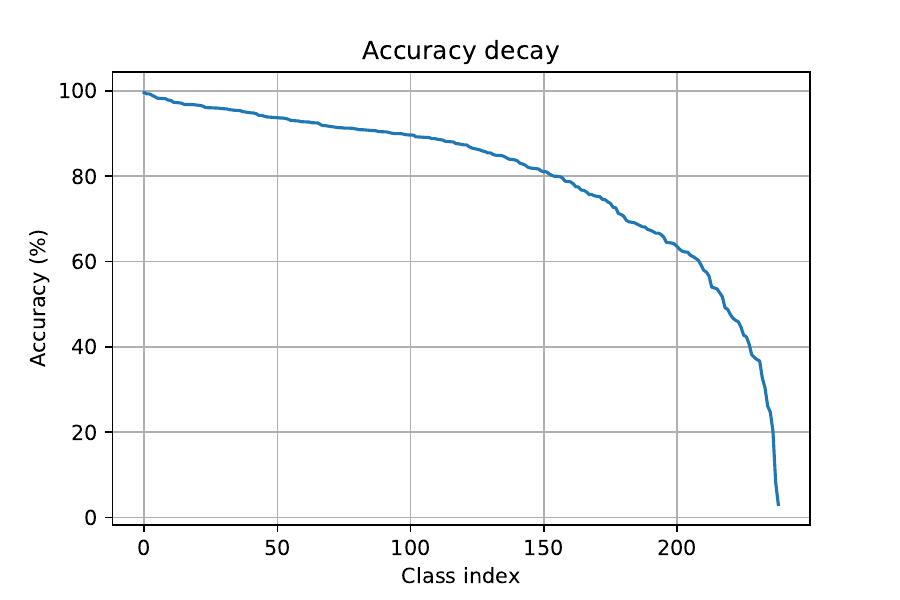}
    \caption{Accuracy decay for FCVID. The chart shows how the accuracy decays after sorting the classes from the highest accuracy to the lowest.}
    \label{fig:accuracy_decay_fcvid}
\end{figure}

To understand if the model behaves well in all classes or it performs very well in some classes and poorly in others we plotted a chart where we show how the accuracy decays from the best performing class to the worst. This chart is shown in Fig.~\ref{fig:accuracy_decay_fcvid}. From the chart we can see that the accuracy is above 80\% for the first 150 classes and it rapidly decays starting at class 200. This means that the model performs very well for most classes and performs poorly for a very small number of classes (around 35 classes have an accuracy below 60\%).
After looking at the classes we found out that the classes where the model performs the best are objects, such as "river", "dog", etc. Most of the classes that perform poorly are in the form making[something], this means that in these classes the model is trying to classify an action and not an object.

\subsubsection{Something-Something}
Our model achieved an accuracy of 21.0\% on the test set of Something-Something which compares with the state-of-the-art of 69.6\% achieved by Swin-B on this dataset. We hypothesise that the transfer-learning did not work properly from Kinetics-400 to Something-Something because the inherent task of the datasets is different. While on Kinetics-400 the objective is to find the main object of the video, on Something-Something the objective is to find relationships between objects and not the object itself.

\begin{figure}[h!]
    \centering
    \includegraphics[width=0.9\linewidth]{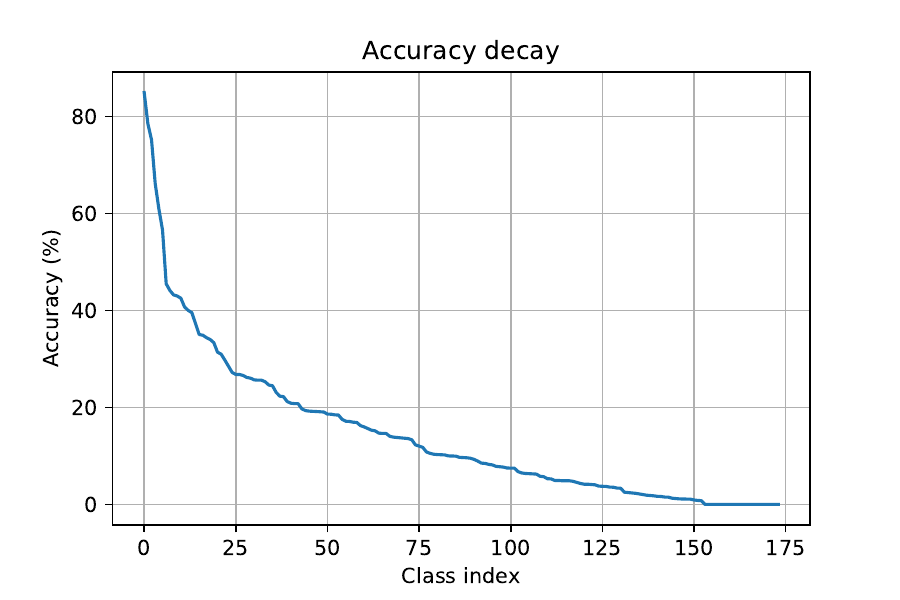}
    \caption{Accuracy decay for Something-Something. The chart shows how the accuracy decays after sorting the classes from the highest accuracy to the lowest. }
    \label{fig:accuracy_decay_ss}
\end{figure}

We plotted a chart where we show how the accuracy decays from the best performing class to the worst. This chart is shown in Fig.~\ref{fig:accuracy_decay_ss}. The accuracy for the first class is above 80\% but it rapidly decays under 30\% and it is close to 0 in the last classes. This means that the model is performs well for a very reduced number of classes and performs poorly for most classes. 

\subsection{Performance Analysis}
To understand the weaknesses of VST we decided to conduct a set analysis on FCVID. 
Given that VST re-scales the videos to 224x224 pixels we hypothesize that the resolution of the original video may impact the performance of the classifier. We also hypothesize that long videos may negatively impact the performance of the classifier because it only takes into account 32 frames per video. This fixed number of frames may not contain enough information to represent the contents of long videos. Finally, we hypothesize that classes underrepresented on the training set may negatively impact the performance of the model. 

To test these three hypotheses we conducted the analysis described below where we assign the videos to a set of bins and then calculate the Spearman's rank correlation between the accuracy and the bins to find out how they correlate. We decided to use Spearman's rank correlation because it evaluates a monotonic relationship instead of a linear relationship, in other words, Spearman's rank correlation evaluates if lower values of one series are paired with lower values of the other series and higher values of one series are paired with higher values of the other series instead of just looking if they are below or above the mean.

\subsubsection{Duration Analysis}
\begin{figure}[h!]
    \centering
    \includegraphics[width=0.9\linewidth]{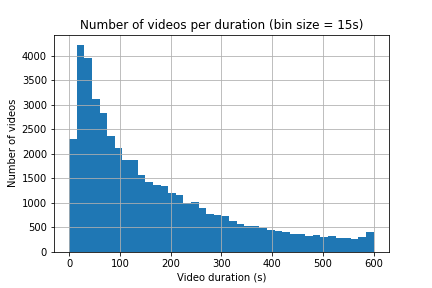}
    \caption{The number of videos from FCVID per bin where each bin has a size of 15 seconds.}
    \label{fig:number_of_videos_per_duration}
\end{figure}

To test if the duration of a video influences the performance of the model, we created bins where a video is assigned to a bin based on its duration. We defined each bin to have a width of 15 seconds. This means that the first bin contains videos with a duration between 0 and 15 seconds, the second bin contains videos with a duration between 15 seconds and 30 seconds, and so on. The size of the bin is a parameter that we chose to create a fair number of bins that we could plot in an image and at the same time have a good amount of videos per bin.
Fig.~\ref{fig:number_of_videos_per_duration} shows the number of videos contained in each bin.

\begin{figure}[h!]
    \centering
    \includegraphics[width=0.9\linewidth]{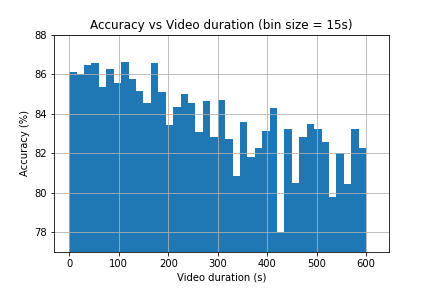}
    \caption{The accuracy per bin, where each bin aggregates videos from FCVID with close duration. Each bin has a width of 15 seconds.}
    \label{fig:accuracy_versus_video_duration}
\end{figure}

Fig.~\ref{fig:accuracy_versus_video_duration} shows the accuracy for each bin. The results show a -0.848 Spearman's rank correlation coefficient which suggests a strong negative correlation between the duration of the video and the accuracy achieved by the model. In other words, this result suggest that the accuracy of the model drops on average when the duration of the videos increases.

\begin{figure}[h!]
    \centering
    \includegraphics[width=0.9\linewidth]{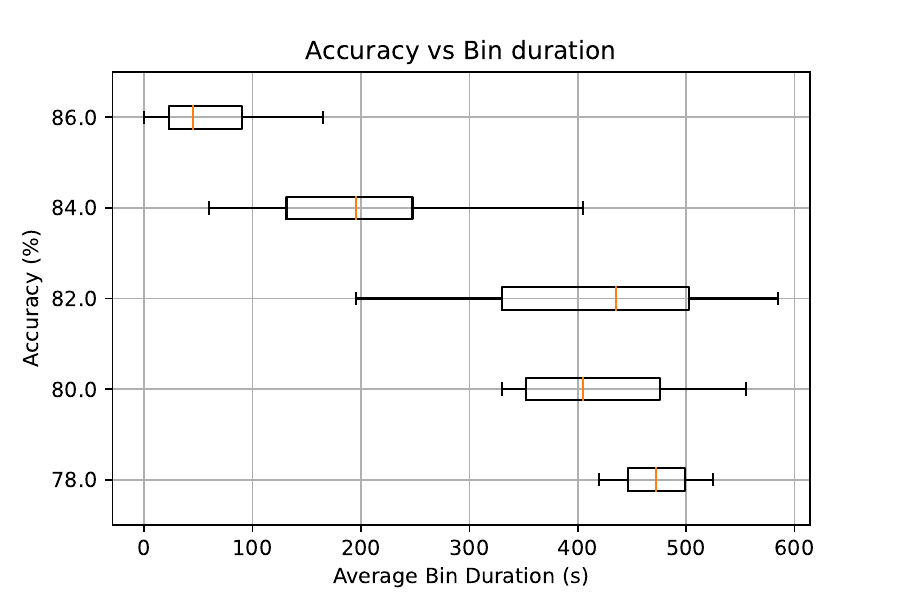}
    \caption{Bin duration versus accuracy for FCVID videos.}
    \label{fig:accuracy_vs_bin_boxplot}
\end{figure}

To further understand how the video duration is related with the accuracy we decided to plot the average bin duration as a function of the accuracy. This plot is shown in Fig.~\ref{fig:accuracy_vs_bin_boxplot}. From the box-plot we can see that bins representing videos with a mean duration of 50s have an accuracy between 86\% to 88\%. On the other side of the plot we can observe that bins with an accuracy between 78\% and 80\% are composed by videos with a mean duration around 480s.

Since the model samples a fixed number of frames from each video, it seems intuitive that those frames are more informative about smaller videos than about longer videos. One reason for this is that the time between each frame in small videos is lesser than in longer videos.

\subsubsection{Resolution Analysis}
\begin{figure}[h!]
    \centering
    \includegraphics[width=0.9\linewidth]{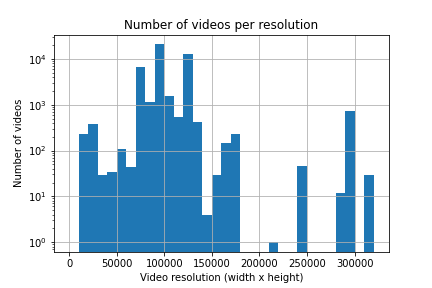}
    \caption{The number of videos from FCVID per bin where each bin has a size of 10,000 pixels. Note that the y axis is on a logarithm scale.}
    \label{fig:number_of_videos_per_resolution}
\end{figure}

Since FCVID scales the input frames to a fixed dimension, one may ask if the resolution and aspect ratio of the original frames may impact the performance of the model.
To test whether or not this is true, we decided to create bins where a video is assigned to a bin based on the total number of pixels per frame (width $\times$ height). We defined each bin to have a width of 10,000 pixels, this means that the first bin contains videos with up to 10,000 pixels per frame, the second bin contains videos with a range of pixels per frame from 10,000 up to 20,000, and so on. Fig.~\ref{fig:number_of_videos_per_resolution} shows the number of videos contained in each bin.

\begin{figure}[h!]
    \centering
    \includegraphics[width=0.9\linewidth]{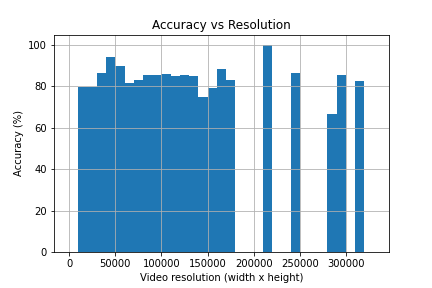}
    \caption{The accuracy per bin, where the bins represent the resolution of the videos from FCVID.}
    \label{fig:accuracy_versus_video_resolution}
\end{figure}

Fig.~\ref{fig:accuracy_versus_video_resolution} shows the accuracy for each bin. The results have a 0.044 Spearman's rank correlation coefficient from which we conclude that the correlation between the resolution of the video and the performance of the model is very weak.
 
\subsubsection{Analysis of the Class Frequency}
Finally, we want to understand if the performance of the model is correlated with the number of samples per class. To test this we calculated the accuracy per class and then calculated the Spearman's rank correlation between the accuracy and the number of samples for each class. We got a 0.334 Spearman's rank correlation which means there is a weak correlation. However, after looking at the classes that have the lowest accuracy we found out that some of these classes can be considered sub-classes of a larger class with more samples. For instance, the class "playingFrisbeeWithDog" has a 3.2\% accuracy and 94 samples, the reason for this is that most of the samples in this class are classified as "dog" which has a 78.4\% accuracy and 399 samples. In other words, when a class can act like a sub-class of another class the model tends to classify the video as belonging to the class with more samples. The same can be observed between "egyptianPyramids" with 3.2\% accuracy and 44 samples and "desert" with 72.6\% accuracy and 226 samples. At the same time we can look at some classes with a low number of samples with high accuracy, for instance, "rockClimbing" with 96.8\% accuracy and 95 samples, "tornado" with 96 samples and 93.8\% accuracy or "playingMahjong" with 63 samples and 93.7\% accuracy. This seems to happen because these classes are very different from all the other classes.

\subsubsection{Analysis of FCVID}
\begin{figure}[h!]
    \centering
    \begin{subfigure}[b]{0.45\linewidth}
        \centering
        \includegraphics[width=\linewidth]{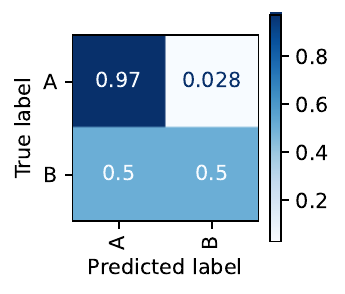}
        \caption{Confusion matrix for the classes soccerAmateur (A) and soccerProfessional (B)}
        \label{fig:confusion_matrix_soccer}
    \end{subfigure}
    \hfill
    \begin{subfigure}[b]{0.45\linewidth}
        \centering
        \includegraphics[width=\linewidth]{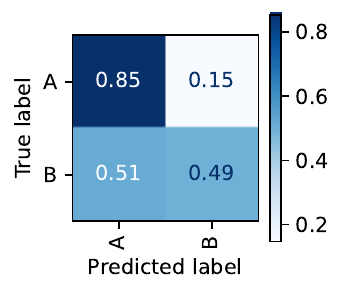}
        \caption{Confusion matrix for the classes basketballAmateur (A) and basketballProfessional (B)}
        \label{fig:confusion_matrix_basketball}
    \end{subfigure}
\end{figure}

We plotted a confusion matrix of the results obtained from FCVID and observed a high confusion between some groups of classes such as
soccerAmateur/soccerProfessional in Fig.~\ref{fig:confusion_matrix_soccer} and basketballAmateur/basketballProfessional in Fig.~\ref{fig:confusion_matrix_basketball}. We attached other examples of high confusion among classes in \ref{appendix_confusion_matrixes}. This confusion suggests that these groups of classes have meanings close to each other. The model in these cases tends to prefer the class with a larger number of samples, for instance, soccerAmateur has 353 samples while soccerProfessional has only 56 samples and so the model tends to classify soccerProfessional as soccerAmateur. The same is true for basketballAmateur with 285 samples and basketballProfessional with 166 samples. If we merged just the mentioned groups the model would increase its accuracy by 1 pp. These results also suggest that there is a hierarchical structure of these classes as we suggested on section \ref{sec:datasets}. This hierarchical structure is however not provided by the creators of the dataset. One example of this are the classes soccerAmateur/soccerProfessional. Soccer can be viewed as a super class while amateur and professional are the subclasses.

\section{Conclusion and future work}
\label{sec:conclusions}
In this paper we aimed to understand if VST generalized well enough to be used in an out of domain setting for video classification without retraining the whole model. The results suggest that VST is capable of extracting features abstract enough that can be used to classify videos of datasets different from the one used for training.
Using this transfer-learning approach, the user is also able to save time and computational resources, since this approach requires around 4x less memory than retraining the whole VST.

The results show that VST generalizes well enough and is applicable to classify out of domain videos without retraining when the target classes are from the same type as the classes used to train the model. In other words, if one trains the model to classify objects, then the model should behave well when classifying objects of never seen classes, we observed this effect when we performed transfer-learning from Kinetics-400 to FCVID, where we got a performance of 85\%. FCVID however contains 30 classes (from 239) which represent actions. These classes have the format making[some action] and the model has an average accuracy of 68\% for these classes which is well below the average of 85\% for the whole dataset.

On the other hand, if the classes are not from the same type, for instance, the training dataset dataset contains objects and the other contains actions, then the accuracy after the transfer-learning approach is expected to be poor. We observed this effect we when performed transfer-learning from Kinetics-400, where the classes are mainly objects to Something-Something, where the classes are mainly actions. We also observed this performance degradation when performing transfer-learning to the classes making[some action] from FCVID.

We suspect this decrease in performance is related to the type of features extracted from the videos, on Kinetics-400 the model learned to identify objects, but to perform well on Something-Something, it would have to learn spacial concepts such as left, right, up, down and the relative position between objects.

We propose as future work to compare the performance of the overlapping classes between FCVID and Kinetics-400 to understand if correlation exists between the accuracy of the same class in different datasets. We also would like to suggest the research and development of zero-shot learning methods for video classification as these would completely eliminate the need for fine-tuning when changing from one domain to another.

\section{Acknowledgments}
Daniel Oliveira is supported by a scholarship granted by Fundação para a Ciência e Tecnologia (FCT), with reference 2021.06750.BD. Additionally, this work was supported by Portuguese national funds through FCT, with reference UIDB/50021/2020.

\bibliographystyle{apalike}
\bibliography{bibliography}

\begin{thebibliography}{}

\bibitem[Arnab et~al., 2021]{vivit}
Arnab, A., Dehghani, M., Heigold, G., Sun, C., Lucic, M., and Schmid, C.
  (2021).
\newblock Vivit: {A} video vision transformer.
\newblock {\em CoRR}, abs/2103.15691.

\bibitem[Bertasius et~al., 2021]{video_understanding}
Bertasius, G., Wang, H., and Torresani, L. (2021).
\newblock Is space-time attention all you need for video understanding?
\newblock {\em CoRR}, abs/2102.05095.

\bibitem[Carreira et~al., 2018]{kinetics_600}
Carreira, J., Noland, E., Banki{-}Horvath, A., Hillier, C., and Zisserman, A.
  (2018).
\newblock A short note about kinetics-600.
\newblock {\em CoRR}, abs/1808.01340.

\bibitem[Carreira et~al., 2019]{kinetics_700}
Carreira, J., Noland, E., Hillier, C., and Zisserman, A. (2019).
\newblock A short note on the kinetics-700 human action dataset.
\newblock {\em CoRR}, abs/1907.06987.

\bibitem[Dosovitskiy et~al., 2020]{vit}
Dosovitskiy, A., Beyer, L., Kolesnikov, A., Weissenborn, D., Zhai, X.,
  Unterthiner, T., Dehghani, M., Minderer, M., Heigold, G., Gelly, S.,
  Uszkoreit, J., and Houlsby, N. (2020).
\newblock An image is worth 16x16 words: Transformers for image recognition at
  scale.
\newblock {\em CoRR}, abs/2010.11929.

\bibitem[Goyal et~al., 2017]{something_something}
Goyal, R., Kahou, S.~E., Michalski, V., Materzynska, J., Westphal, S., Kim, H.,
  Haenel, V., Fr{\"{u}}nd, I., Yianilos, P., Mueller{-}Freitag, M., Hoppe, F.,
  Thurau, C., Bax, I., and Memisevic, R. (2017).
\newblock The "something something" video database for learning and evaluating
  visual common sense.
\newblock {\em CoRR}, abs/1706.04261.

\bibitem[Jiang et~al., 2016]{fcvid}
Jiang, Y.-G., Wu, Z., Wang, J., Xue, X., and Chang, S.-F. (2016).
\newblock Fcvid : Fudan-columbia video dataset.

\bibitem[Kay et~al., 2017]{kinetics_400}
Kay, W., Carreira, J., Simonyan, K., Zhang, B., Hillier, C., Vijayanarasimhan,
  S., Viola, F., Green, T., Back, T., Natsev, P., Suleyman, M., and Zisserman,
  A. (2017).
\newblock The kinetics human action video dataset.
\newblock {\em CoRR}, abs/1705.06950.

\bibitem[Liu et~al., 2021a]{swin_transformer}
Liu, Z., Lin, Y., Cao, Y., Hu, H., Wei, Y., Zhang, Z., Lin, S., and Guo, B.
  (2021a).
\newblock Swin transformer: Hierarchical vision transformer using shifted
  windows.
\newblock {\em CoRR}, abs/2103.14030.

\bibitem[Liu et~al., 2021b]{video_swin_transformer}
Liu, Z., Ning, J., Cao, Y., Wei, Y., Zhang, Z., Lin, S., and Hu, H. (2021b).
\newblock Video swin transformer.
\newblock {\em arXiv preprint arXiv:2106.13230}.

\bibitem[Loshchilov and Hutter, 2017]{adamw}
Loshchilov, I. and Hutter, F. (2017).
\newblock Decoupled weight decay regularization.
\newblock {\em arXiv preprint arXiv:1711.05101}.

\bibitem[Tran et~al., 2014]{3d_convolution}
Tran, D., Bourdev, L.~D., Fergus, R., Torresani, L., and Paluri, M. (2014).
\newblock {C3D:} generic features for video analysis.
\newblock {\em CoRR}, abs/1412.0767.

\end{thebibliography}

\appendix
\section{High confusion among some groups of classes}
\label{appendix_confusion_matrixes}

\begin{figure}[h!]
    \centering
    \begin{subfigure}[b]{0.45\linewidth}
        \centering
        \includegraphics[width=\linewidth]{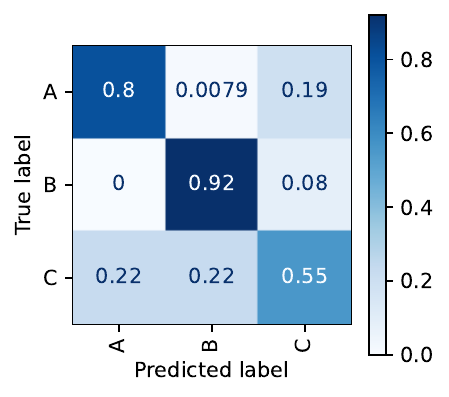}
        \caption{Confusion matrix for the classes weddingDance (A), weddingCeremony (B) and weddingReception (C)}
        \label{fig:confusion_matrix_wedding}
    \end{subfigure}
    \hfill
    \begin{subfigure}[b]{0.45\linewidth}
        \centering
        \includegraphics[width=\linewidth]{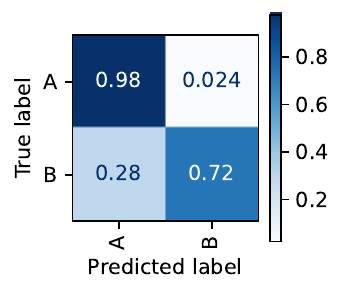}
        \caption{Confusion matrix for the classes swimmingAmateur (A) and swimmingProfessional (B)}
        \label{fig:confusion_matrix_swimming}
    \end{subfigure}
\end{figure}

\begin{figure}[h!]
    \centering
    \begin{subfigure}[b]{0.45\linewidth}
        \centering
        \includegraphics[width=\linewidth]{swimming.pdf}
        \caption{Confusion matrix for the classes swimmingAmateur (A) and swimmingProfessional (B)}
        \label{fig:confusion_matrix_swimming}
    \end{subfigure}
    \hfill
    \begin{subfigure}[b]{0.45\linewidth}
        \centering
        \includegraphics[width=\linewidth]{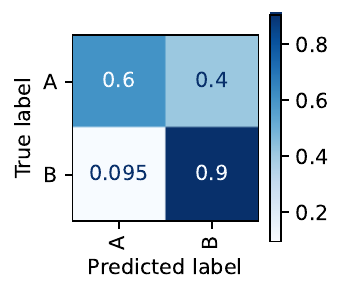}
        \caption{Confusion matrix for the classes townHallMetting (A), publicSpeech (A)}
        \label{fig:confusion_matrix_public_speech}
    \end{subfigure}
\end{figure}

\begin{figure}[h!]
    \centering
    \begin{subfigure}[b]{0.45\linewidth}
        \centering
        \includegraphics[width=\linewidth]{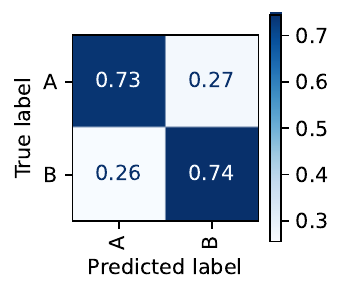}
        \caption{Confusion matrix for the classes americanFootballAmateur (A), americanFootballProfessional (B)}
        \label{fig:confusion_matrix_american_football}
    \end{subfigure}
    \hfill
    \begin{subfigure}[b]{0.45\linewidth}
        \centering
        \includegraphics[width=\linewidth]{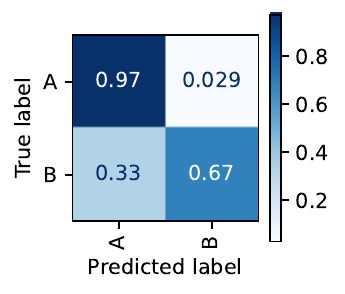}
    \caption{Confusion matrix for the classes diningAtRestaurant (A), dinnerAtHome (B)}
    \label{fig:confusion_matrix_dining}
    \end{subfigure}
\end{figure}

\end{document}